\documentclass[10pt]{article}
\usepackage[preprint]{tmlr}

\usepackage{amsmath,amsfonts,bm}

\def\eqref#1{equation~\ref{#1}}

\def\1{\bm{1}}

\DeclareMathAlphabet{\mathsfit}{\encodingdefault}{\sfdefault}{m}{sl}
\SetMathAlphabet{\mathsfit}{bold}{\encodingdefault}{\sfdefault}{bx}{n}

\usepackage{hyperref}
\usepackage{url}
\usepackage{graphicx}
\usepackage{booktabs}
\usepackage{amsmath,amssymb}
\usepackage{tikz}
\usepackage{float}
\usetikzlibrary{arrows.meta, positioning}

\title{Recurrent Temporal Restricted Boltzmann Machines as Unsupervised Encoders for Biomechanical Movement Sequences}

\author{\name Wafi Sharif \\ \addr Queens High School for the Sciences \AND \name Nicholas LaHaye \\ \addr Jet Propulsion Laboratory, California Institute of Technology}

\def\month{MM}
\def\year{YYYY}
\def\openreview{\url{https://openreview.net/forum?id=XXXX}}

\begin{document}

\maketitle

\begin{abstract}
Biomechanical movement recordings are continuous, high-dimensional time series, and analyzing them without manually labeling them requires representations that capture how movements evolve over time. Most unsupervised representation-learning pipelines, however, treat each observation frame independently and discard this temporal structure, while energy-based models that could capture it have not extensively been adapted to biomechanical signals. So, this paper implements the Recurrent Temporal Restricted Boltzmann Machine (RTRBM), a recurrent energy-based model, into an unsupervised encoder that maps windows of movement data to fixed-length embeddings, and tests what its recurrent structure contributes. We adapt the model to continuous-valued signals, stack it into a deep architecture, and integrate it into the SIT-FUSE segmentation pipeline. We evaluate the embeddings on simulated musculoskeletal reaching and on two real gait datasets, using clustering, reconstruction, and walking-speed classification of unseen subjects, and compare them against non-recurrent and non-probabilistic baselines, including a control that removes the recurrent connection. Clustering the embeddings recovers movement structure tied to progress toward the target without any labels, which also transfers to unseen episodes. On gait data, the embeddings outperform PCA and autoencoder baselines under a nonlinear probe. However, removing the recurrent connection does not yield a measurable loss on the reaching task, and simple per-channel summary statistics classify walking speed at least as well as the embeddings. The model is therefore a working, verified encoder for SIT-FUSE, while the evidence for a benefit of recurrence is mixed, and the learned embeddings did not outperform simple summary statistics on gait.
\end{abstract}

\section{Introduction}

Biomechanical time series are high-dimensional, continuous, and inherently sequential. As features like joint kinematics and contact forces evolve in ways that reflect an underlying motor plan, significant information in a movement is spread across time. Therefore, treating each observation as an independent sample takes away that structure and reduces a movement to simply a set of disconnected snapshots.

The typical fix for this has been supervision, by training a classifier to predict movement phases from annotated recordings \citep{altimemy2022deeplearningmovement}. However, exhaustively hand-labeling recordings is costly, and models trained on labeled recordings often degrade when the sensor placement, device, or subject population changes \citep{chakma2023domain,dhekane2024transfer}. Meanwhile, unsupervised methods that can directly find temporal structure from raw data alleviate the effort required to produce labels, instead shifting efforts towards interpreting the structure a model finds. This overall motivates the use of unsupervised methods here.

Restricted Boltzmann Machines (RBMs) are undirected energy-based models used in unsupervised representation learning \citep{hinton2006fast}; stacking them into Deep Belief Networks (DBNs) trained layer-by-layer helps to pull out abstract features \citep{hinton2006fast}. Feeding these hidden representations into downstream clustering is the approach SIT-FUSE \citep{lahaye2021sitfuse} takes for self-supervised segmentation of remote sensing imagery; the code here is a sequence-aware extension of that pipeline, and we use biomechanical reaching trajectories to validate it. Every standard RBM processes each input frame by itself. Nothing conditions the hidden state at time $t$ on what the model saw at $t{-}1$, so no temporal context carries forward.

\begin{figure}[t]
\centering
\resizebox{\linewidth}{!}{%
\begin{tikzpicture}[
    >=Latex,
    vis/.style={circle, draw, minimum size=9mm, fill=blue!12, inner sep=0pt, font=\tiny},
    hid/.style={circle, draw, minimum size=9mm, fill=orange!20, inner sep=0pt, font=\tiny},
    rnn/.style={circle, draw, minimum size=9mm, fill=green!18, inner sep=0pt, font=\tiny},
    lbl/.style={font=\scriptsize},
    panel/.style={font=\small}
]

\begin{scope}
\node[hid] (a-h) at (0,1.1) {$h$};
\node[vis] (a-v) at (0,0) {$v$};
\draw[<->] (a-v) -- node[right, lbl] {$W$} (a-h);
\node[panel] at (0,-0.9) {(a) RBM};
\end{scope}

\begin{scope}[xshift=4.4cm]
\node[hid] (b-h1) at (0,1.1) {$h_{t-1}$};
\node[vis] (b-v1) at (0,0) {$v_{t-1}$};
\node[hid] (b-h2) at (2.1,1.1) {$h_{t}$};
\node[vis] (b-v2) at (2.1,0) {$v_{t}$};
\node[hid] (b-h3) at (4.2,1.1) {$h_{t+1}$};
\node[vis] (b-v3) at (4.2,0) {$v_{t+1}$};

\draw[<->] (b-v1) -- node[left, lbl] {$W$} (b-h1);
\draw[<->] (b-v2) -- node[left, lbl] {$W$} (b-h2);
\draw[<->] (b-v3) -- node[left, lbl] {$W$} (b-h3);

\draw[->, thick, red!70!black] (b-h1) -- node[above, lbl] {$W'$} (b-h2);
\draw[->, thick, red!70!black] (b-h2) -- node[above, lbl] {$W'$} (b-h3);

\node[panel] at (2.1,-0.9) {(b) RTRBM (this work)};
\end{scope}

\begin{scope}[xshift=11.6cm]
\node[hid] (c-h1) at (0,1.1) {$h_{t-1}$};
\node[vis] (c-v1) at (0,0) {$v_{t-1}$};
\node[rnn] (c-r1) at (0,2.5) {$r_{t-1}$};

\node[hid] (c-h2) at (2.4,1.1) {$h_{t}$};
\node[vis] (c-v2) at (2.4,0) {$v_{t}$};
\node[rnn] (c-r2) at (2.4,2.5) {$r_{t}$};

\node[hid] (c-h3) at (4.8,1.1) {$h_{t+1}$};
\node[vis] (c-v3) at (4.8,0) {$v_{t+1}$};
\node[rnn] (c-r3) at (4.8,2.5) {$r_{t+1}$};

\draw[<->] (c-v1) -- node[left, lbl] {$W$} (c-h1);
\draw[<->] (c-v2) -- node[left, lbl] {$W$} (c-h2);
\draw[<->] (c-v3) -- node[left, lbl] {$W$} (c-h3);

\draw[->, thick, teal!70!black] (c-r1) -- (c-r2);
\draw[->, thick, teal!70!black] (c-r2) -- (c-r3);

\draw[->, thick, teal!70!black] (c-v1) to[bend right=18] (c-r1);
\draw[->, thick, teal!70!black] (c-v2) to[bend right=18] (c-r2);
\draw[->, thick, teal!70!black] (c-v3) to[bend right=18] (c-r3);

\draw[->, thick, teal!70!black] (c-r1) -- (c-h2);
\draw[->, thick, teal!70!black] (c-r2) -- (c-h3);

\node[panel] at (2.4,-0.9) {(c) RNN-RBM};
\end{scope}

\end{tikzpicture}%
}
\caption{Three ways to add memory to an RBM. (a) A standard RBM has no temporal connection; each frame is modeled independently. (b) The RTRBM used in this work reuses its own hidden units as the recurrent signal: $h_{t-1}$ sets the hidden bias at $t$ through a single learned weight matrix $W'$ (red). (c) The RNN-RBM keeps a separate recurrent state $r_t$, updated from the current frame and the previous $r$, which then sets the RBM's bias one step ahead (teal): more expressive, but with a second set of parameters to train and coordinate.}
\label{fig:architectures}
\end{figure}
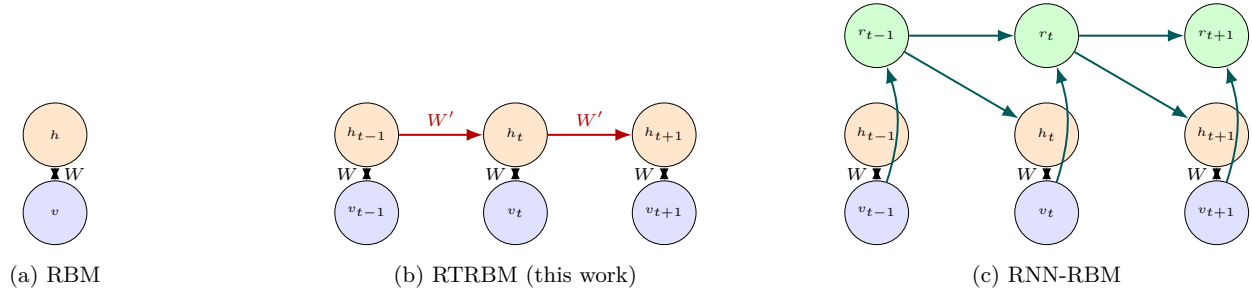

The Recurrent Temporal Restricted Boltzmann Machine \citep{sutskever2008recurrent} closes that gap (Figure~\ref{fig:architectures}). It augments the RBM with a learned hidden-to-hidden weight matrix $\mathbf{W}'$ that carries mean-field hidden probabilities forward in time, modulating the hidden bias at each step. It defines a generative distribution over whole sequences, and the original RTRBM paper showed that it generates more realistic motion-capture sequences and bouncing-ball videos than the temporal RBM (TRBM) it modifies \citep{sutskever2008recurrent}. Its motion-capture experiments used Gaussian visible units with a fixed unit variance. Biomechanical recordings combine joint angles, velocities, and muscle activations whose variability differs from channel to channel, so we extend the model with a learned per-feature variance, which requires care in how the variance is trained \citep{cho2011improved} and improves held-out reconstruction (\S\ref{sec:modelcomp}). Temporal RBMs have so far been applied to human movement mainly to generate, predict, or fill in motion-capture sequences \citep{taylor2011two,sukhbaatar2011robust,sutskever2008recurrent}. To our knowledge, they have not been used as unsupervised encoders that turn windows of biomechanical signals into fixed-length embeddings for clustering, which is the role we develop here.

A related model, known as the RNN-RBM \citep{boulanger2012modeling}, conditions each RBM in the sequence on the hidden state of an RNN instead of on a direct hidden-to-hidden connection. This gives it more expressive power, but also makes it harder to train and requires separate optimization of the RNN and RBM components. The RTRBM's single recurrent matrix is a simpler inductive bias that suits the short 20-step windows used here.

This paper describes how we developed the RTRBM into a working temporal encoder for SIT-FUSE. We implement the Bernoulli RTRBM and a Gaussian-visible variant with a learned per-feature variance \citep{cho2011improved}, trained with a two-phase variance schedule, gradient clipping, and a sampled negative particle that keep training stable on continuous multichannel signals. We stack RTRBM layers into a Recurrent Temporal Deep Belief Network (RTDBN) that turns each movement window into one fixed-length embedding by temporal mean pooling, with windowing that never spans two recordings. The implementation follows Learnergy's conventions, and its training gradients agree with exact maximum likelihood on small enumerable models. Our work is also integrated into SIT-FUSE as a selectable encoder. We evaluate it on simulated reaching trajectories from MyoSuite \citep{myosuite2022}, where $k$-means clusters of embeddings are organized around target distance and closing rate, as well as on two real gait cohorts, where the embeddings predict walking speed in unseen subjects better than PCA and autoencoder embeddings under a random-forest probe, but not better than simple per-channel summary statistics. Further, an ablation study finds no benefit from the recurrent connection itself on the reach data (\S\ref{sec:wprime}).

\section{Methods}

\subsection{Restricted Boltzmann Machines}

An RBM defines a joint distribution over visible units $\mathbf{v} \in \mathbb{R}^{n_v}$ and binary hidden units $\mathbf{h} \in \{0,1\}^{n_h}$ through the energy:
\begin{equation}
E(\mathbf{v}, \mathbf{h}) =
-\mathbf{v}^{\top}\mathbf{a}
- \mathbf{h}^{\top}\mathbf{b}
- \mathbf{v}^{\top}\mathbf{W}\mathbf{h}
\label{eq:rbm_energy}
\end{equation}
where $\mathbf{a}$, $\mathbf{b}$ are bias vectors and $\mathbf{W}$ is the weight matrix. The bipartite structure makes block Gibbs sampling tractable, and parameters are learned by Contrastive Divergence \citep{hinton2002training}, which runs a $k$-step Gibbs chain from a data point to approximate the model expectation.

\subsection{The Recurrent Temporal RBM}

The RTRBM \citep{sutskever2008recurrent} adds a recurrent weight matrix $\mathbf{W}' \in \mathbb{R}^{n_h \times n_h}$ and a learned initial hidden state $\mathbf{h}_0$. For a sequence $\{\mathbf{v}_1, \ldots, \mathbf{v}_T\}$, the hidden bias at step $t$ becomes:
\begin{equation}
\tilde{\mathbf{b}}_t = \mathbf{b} + \mathbf{W}'\hat{\mathbf{h}}_{t-1}
\label{eq:rtrbm_bias}
\end{equation}
where $\hat{\mathbf{h}}_{t-1}$ are the mean-field hidden probabilities from the previous step. The per-step conditionals are:
\begin{align}
p(h_{t,j} = 1 \mid \mathbf{v}_t, \hat{\mathbf{h}}_{t-1}) &=
\sigma\!\left(\tilde{b}_{t,j} + \sum_i W_{ij}\,v_{t,i}\right)
\label{eq:rtrbm_hidden} \\[4pt]
p(v_{t,i} \mid \mathbf{h}_t) &=
\sigma\!\left(a_i + \sum_j W_{ij}\,h_{t,j}\right)
\quad \text{(Bernoulli)}
\label{eq:rtrbm_visible}
\end{align}

\subsection{Gaussian Visible Units and Learned Variance}

Binary visible outputs do not fit continuous biomechanical signals. Swapping the Bernoulli layer for Gaussian units changes the energy to:
\begin{equation}
E(\mathbf{v}, \mathbf{h}) =
\sum_i \frac{(v_i - a_i)^2}{2\sigma_i^2}
- \sum_j \tilde{b}_{t,j}\,h_j
- \sum_{i,j} \frac{W_{ij}\,v_i\,h_j}{\sigma_i^2}
\label{eq:gaussian_energy}
\end{equation}
with visible conditional $p(v_i \mid \mathbf{h}) = \mathcal{N}(a_i + \sum_j W_{ij} h_j,\; \sigma_i^2)$ and hidden conditional:
\begin{equation}
p(h_j = 1 \mid \mathbf{v}) =
\sigma\!\left(\tilde{b}_j +
\sum_i \frac{W_{ij}\,v_i}{\sigma_i^2}\right)
\label{eq:gaussian_hidden}
\end{equation}
Making $\boldsymbol{\sigma}$ a learnable per-feature parameter lets each channel's output scale adapt to its own variance. This is the parameterization from \citet{cho2011improved}, which we call RTVarianceGaussianRBM.

\subsection{Training RTVarianceGaussianRBM}
\label{sec:training}

We follow the two-phase schedule of \citet{cho2011improved}: $\boldsymbol{\sigma}$ is frozen at 1 for 15 epochs while $\mathbf{W}$, $\mathbf{W}'$, $\mathbf{a}$, $\mathbf{b}$, and $\mathbf{h}_0$ stabilize, then unfrozen for 15 more. We also clip gradients (global max norm $1.0$) and clamp $\sigma_i \in [0.1, 10.0]$ after every optimizer step.

$\boldsymbol{\sigma}$ is optimized directly as a standard deviation. $\mathbf{W}$ and $\mathbf{W}'$ are initialized from $\mathcal{N}(0, 0.01^2)$; all bias vectors ($\mathbf{a}$, $\mathbf{b}$, $\mathbf{h}_0$) are initialized to zero; $\boldsymbol{\sigma}$ is initialized to $\mathbf{1}$. We use plain SGD, with batch size 32, on CPU. We selected each model's learning rate by a grid search on validation reconstruction MSE, which was 1.0 for the canonical and $\mathbf{W}'=0$ models, 0.3 for the fixed-variance Gaussian model, and 0.1 for the Bernoulli model. In the canonical reach model, the learning rate of the group containing $\mathbf{W}'$ and $\mathbf{h}_0$ is multiplied by 0.01; on the gait data this multiplier is 1, so all weights share one learning rate. Unless stated otherwise, every result below is over ten seeds (1--10); the seed controls the 70/15/15 episode split, weight initialization, and $k$-means initialization, and models are compared paired by seed (same split). We report mean $\pm$ SD, paired differences with 95\% bootstrap confidence intervals.

CD-1 is performed independently at each timestep, conditioned on the previous step's mean-field hidden probability $\hat{\mathbf{h}}_{t-1}$. Within a single timestep's Gibbs step, the visible unit is sampled from its Gaussian conditional; this sample is the negative particle in the energy term, while the reconstruction-quality metric uses the conditional mean.

\begin{verbatim}
Algorithm: CD-1 training step for RTVarianceGaussianRBM, one window
Input: window v_{1..T} in R^{T x n_v}, previous states h_0
  h_prev <- h_0 (broadcast to batch)
  total_cost <- 0
  for t = 1 to T:
    // positive phase (teacher-forced on real data)
    pos_probs, pos_states <- hidden_sampling(v_t, h_prev)
    // negative phase (CD-1, one Gibbs step)
    neg_h <- pos_states
    v_recon, v_recon_mean <- visible_sampling(neg_h) // v_recon ~ N(mean, sigma^2)
    neg_probs, neg_h <- hidden_sampling(v_recon, h_prev)
    v_recon <- detach(v_recon)                  // stop-gradient
    cost_t <- energy(v_t, h_prev) - energy(v_recon, h_prev)
    mse_t <- sum((v_t - v_recon_mean)^2) / batch_size // metric only (mean)
    total_cost <- total_cost + cost_t
    // advance recurrent state using REAL data (teacher forcing), not reconstruction
    h_prev <- hidden_sampling(v_t, h_prev).probs
  total_cost.backward()
  clip_grad_norm(params, max_norm=1.0)
  optimizer.step()
  sigma <- clamp(sigma, 0.1, 10.0)
\end{verbatim}

On the reach data, we train on overlapping windows of length $T=20$ with stride 1 to keep BPTT tractable and yield more training samples; the gait data use non-overlapping windows (\S\ref{sec:gaitdata}).

\subsection{Recurrent Temporal Deep Belief Network}

\begin{figure}[t]
\centering
\begin{tikzpicture}[
    >=Latex,
    seqnode/.style={circle, draw, minimum size=6mm, fill=blue!12, inner sep=0pt, font=\tiny},
    box/.style={draw, rounded corners, minimum height=9mm, minimum width=20mm, align=center, font=\scriptsize, fill=orange!12},
    embnode/.style={rectangle, draw, minimum height=9mm, minimum width=9mm, fill=purple!12, font=\tiny},
    lbl/.style={font=\scriptsize}
]

\node[seqnode] (v1) at (0,0) {$v_1$};
\node[seqnode] (v2) at (0.9,0) {$v_2$};
\node at (1.65,0) {\scriptsize $\cdots$};
\node[seqnode] (vT) at (2.4,0) {$v_T$};
\node[lbl, above=1mm of v1, anchor=south west, xshift=-2mm] {input window, length $T$};

\node[box, minimum width=32mm] (rtrbm) at (1.2,-1.6) {RTRBM layer\\ $h_t \gets W, W', h_{t-1}$};
\draw[->] (v1) -- (rtrbm.north -| v1);
\draw[->] (v2) -- (rtrbm.north -| v2);
\draw[->] (vT) -- (rtrbm.north -| vT);

\node[seqnode, fill=orange!25] (h1) at (0,-3.2) {$h_1$};
\node[seqnode, fill=orange!25] (h2) at (0.9,-3.2) {$h_2$};
\node at (1.65,-3.2) {\scriptsize $\cdots$};
\node[seqnode, fill=orange!25] (hT) at (2.4,-3.2) {$h_T$};
\draw[->] (rtrbm.south -| h1) -- (h1);
\draw[->] (rtrbm.south -| h2) -- (h2);
\draw[->] (rtrbm.south -| hT) -- (hT);

\node[box, minimum width=32mm, fill=teal!12] (pool) at (1.2,-4.8) {mean pool over $t$\\ $\mathbf{e} = \frac{1}{T}\sum_t \hat{\mathbf{h}}_t$};
\draw[->] (h1) -- (pool.north -| h1);
\draw[->] (h2) -- (pool.north -| h2);
\draw[->] (hT) -- (pool.north -| hT);

\node[embnode, minimum width=14mm, minimum height=7mm] (emb) at (1.2,-6.2) {$\mathbf{e} \in \mathbb{R}^{64}$};
\draw[->] (pool) -- (emb);

\node[box, minimum width=32mm, fill=green!12] (km) at (1.2,-7.6) {$k$-means, $k = 10$};
\draw[->] (emb) -- (km);

\node[lbl] (out) at (1.2,-8.7) {movement-phase cluster};
\draw[->] (km) -- (out);

\end{tikzpicture}
\caption{The full RTDBN pipeline for a single window of length $T$: each frame $v_t$ passes through the RTRBM layer, whose recurrent connection carries information from preceding frames; then, the resulting hidden sequence is mean-pooled into one embedding per window, which $k$-means then assigns to a movement-phase cluster.}
\label{fig:pipeline}
\end{figure}
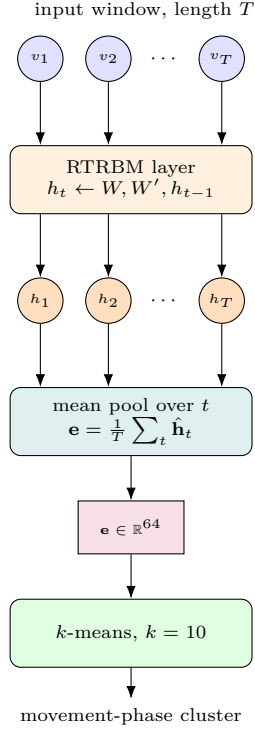

The RTDBN stacks RTRBM layers using greedy layer-wise pretraining \citep{hinton2006fast}. After the final layer, we collapse the temporal hidden states into a single embedding per window by mean pooling (Figure~\ref{fig:pipeline}):
\begin{equation}
\mathbf{e} = \frac{1}{T}\sum_{t=1}^T \hat{\mathbf{h}}_t
\label{eq:mean_pool}
\end{equation}
All experiments here use a single RTRBM layer; stacking more is implemented but unnecessary at this data scale.

\subsection{Clustering}

We cluster the embeddings $\{\mathbf{e}_i\}$ with $k$-means ($k = 10$, 20 random restarts), instead of the Invariant Information Clustering (IIC) loss \citep{ji2019invariant} that SIT-FUSE's full pipeline normally uses. We use $k$-means as a deliberately decoupled clustering readout, separate from encoder training.

We also report silhouette score \citep{rousseeuw1987silhouettes}, which measures how much closer each point is to points in its own cluster than the points of other clusters, and normalized cluster entropy:
\begin{equation}
H_{\text{norm}} = -\frac{1}{\log k}\sum_{c=1}^k
\frac{n_c}{N}\log\frac{n_c}{N}
\label{eq:entropy}
\end{equation}
$H_{\text{norm}} = 1$ is uniform cluster sizes; $H_{\text{norm}} = 0$ is complete collapse. Because silhouette depends on embedding geometry, we also report (i) held-out per-channel $R^2 = 1 - \text{MSE}_c/\text{Var}_c$ of the mean reconstruction (reconstruction from sampled hidden states), summarized by its median over channels and by channel group (positions 0--19, velocities 20--39, tip position 40--42, reach error 43--45, muscle activations 46--79), and (ii) ridge-regression probes (penalty 1.0) fit on the train+validation embeddings and scored by $R^2$ on the test embeddings, for the following targets: the change in reach error over the window (reach trend), mean joint speed, and the relative position of the window within its episode (phase). We also report normalized mutual information (NMI) between the $k$-means clusters and ten equal phase bins.

\subsection{Data}

We generate reaching data from the \texttt{myoArmReachFixed-v0} environment in MyoSuite \citep{myosuite2022}, which uses MuJoCo (Multi-Joint dynamics with Contact) \citep{todorov2012mujoco} to simulate a physiological upper-limb model on a fixed-target reaching task. We let 100 episodes run for 151 steps each, with an 80-dimensional observation vector: joint positions (20), joint velocities (20), reach-site position (3), reach error (displacement between target and reach site) (3), and muscle activations (34).

Actions are drawn from a momentum-biased random policy. At each timestep, an action is drawn from a Gaussian centered on a slowly-adapting drift vector; if the previous step reduced the distance to the target, the drift vector is nudged toward the action that produced the improvement, but it decays back toward zero if not.

We partition into 70 training, 15 validation, and 15 test set episodes, ensuring that overlapping windows from the same episode do not cross a split boundary. A StandardScaler (zero-mean, unit-variance per channel) is fit on training episodes and applied to validation and test data at inference time.

We extend SIT-FUSE's data pipeline to handle multi-trial biomechanical recordings. A causal sliding window of length $T$ is applied within each episode before pooling.

\subsection{Software Implementation, Verification and Integration with SIT-FUSE}
\label{sec:software}

\paragraph{Library.} The temporal models are implemented as Learnergy classes: \texttt{RTRBM} (Bernoulli), \texttt{RTGaussianRBM} (fixed variance), \texttt{RTVarianceGaussianRBM} (learned per-feature variance), and \texttt{RTDBN}, which stacks RTRBM layers greedily.

\paragraph{Verification.} We checked the training gradients against exact maximum likelihood on enumerable models. With the exact negative-phase expectation, the contrastive-divergence surrogate used by the code equals the automatic-differentiation gradient of the exact sequence log-likelihood, taken through the recurrent chain of the learned-variance model. The conditionals implied by the joint energy equal those the classes sample from.

\paragraph{Integration with SIT-FUSE.} Setting \texttt{encoder\_type: rtdbn} in a SIT-FUSE configuration file selects the temporal encoder. The stages are: encoder pretraining, the clustering stage, inference that writes one cluster label per window for the held-out splits, FLOP counting, and per-cluster TimeSHAP attributions over time steps and channels.

\subsection{Real Gait Data and Probe Protocol}
\label{sec:gaitdata}

To test the embeddings outside simulation, we use two independent human gait cohorts where the task is to classify walking speed (three classes, chance $= 0.333$) from joint-angle windows of subjects that were not used for training. Cohort~1 contains 57 subjects walking at three speeds (0.5, 0.75 and 1.0\,m/s), with 66 inertial-sensor-derived joint-angle channels at 60\,Hz \citep{katmah2026multimodal,pollard2026physionet}. Cohort~2 is the dataset of \citet{bailey2021validity}, released through SimTK \citep{bailey2021simtk}: 14 healthy adults walking on a treadmill at 70\%, 100\%, and 130\% of their preferred speed, with IMU-driven joint angles from inverse kinematics; we keep the 18 rotational channels that vary by more than one degree (a label-free rule) and drop the first five seconds of each trial.

Windows are non-overlapping with $T=30$ frames, standardized with training-subject statistics. Splits are at the subject level (70/15/15 for cohort~1; 10/2/2 subjects for cohort~2), with ten splits per cohort. Each model produces a 64-dimensional embedding (the RTRBM variants by temporal mean pooling), trained for 30 epochs with the schedule of \S\ref{sec:training} and a learning rate chosen only from validation reconstruction error (0.1 for cohort~1, 0.3 for cohort~2). The comparison encoders are frame-wise PCA, window PCA, an MLP autoencoder, and a GRU autoencoder, all with 64-dimensional embeddings, and a training-free baseline whose embedding is the per-channel mean, standard deviation, minimum and maximum over the window ($4\times$ the number of channels: 264 features for cohort~1 and 72 for cohort~2). Two probes are fit on frozen embeddings and scored by test accuracy on unseen subjects: logistic regression and a random forest (40 trees). Contrasts are paired over the ten splits with 95\% bootstrap intervals and Holm correction.

\section{Results}

\subsection{Model Comparison}
\label{sec:modelcomp}

\begin{figure}[h]
\begin{center}
\includegraphics[width=\linewidth]{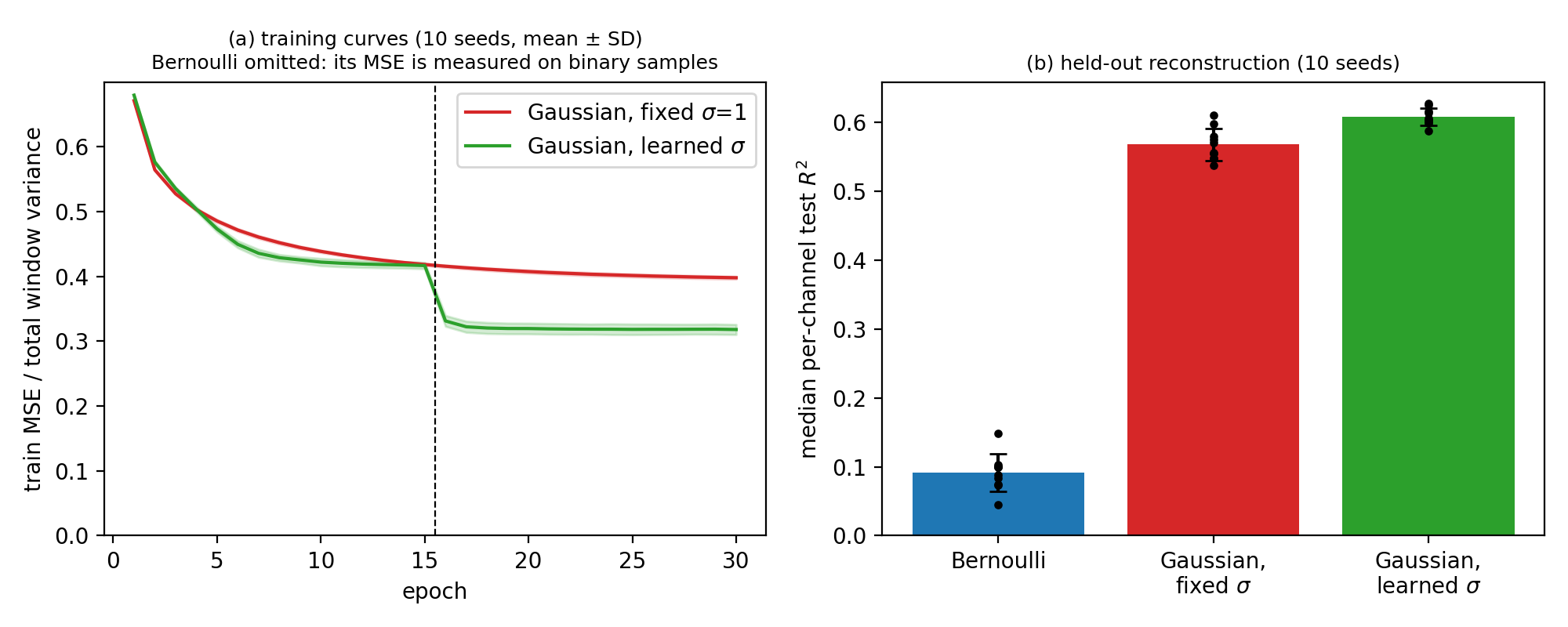}
\end{center}
\caption{
(a) Training MSE divided by total window variance (mean $\pm$ SD over ten seeds) for the fixed-variance Gaussian (red) and learned-variance Gaussian (green) RTRBMs; the dashed line marks epoch 15, when $\sigma$ is unfrozen. The Bernoulli model is omitted from (a) because its MSE is measured against binary samples. (b) Median per-channel held-out $R^2$ for all three variants (bars: mean $\pm$ SD; dots: individual seeds).
}
\label{fig:mse_comparison}
\end{figure}

We run three RTRBM variants at $n_v = 80$, $n_h = 64$, CD-1, $T = 20$, for 30 epochs and ten seeds, changing the visible unit parameterization: Bernoulli units on min--max scaled data, and fixed-variance and learned-variance Gaussian units on standardized data. Aggregate MSE is not comparable across these (the Bernoulli model's MSE is computed against binary samples and its data are on a different scale), so we compare held-out per-channel $R^2$ computed from the mean reconstruction. The Bernoulli model reaches a median $R^2$ of $0.09 \pm 0.03$, the fixed-variance Gaussian model $0.57 \pm 0.02$, and RTVarianceGaussianRBM $0.61 \pm 0.01$. The paired difference between the learned- and fixed-variance models is $+0.04$ (95\% CI $[+0.025, +0.056]$) in median $R^2$ and $-189$ in held-out MSE ($539 \pm 27$ versus $728 \pm 21$; CI $[-199, -177]$). With the learned variance, joint kinematics reconstruct well (median $R^2$ 0.86 for positions and 0.76 for velocities) and muscle activations moderately (0.56) (Figure~\ref{fig:mse_comparison}).

\subsection{RTDBN Embeddings and Movement Clustering}

Applying the single-layer RTDBN ($n_h = 64$, $k = 10$) to embeddings from the frozen encoder gives 11{,}220 train+validation windows split across 10 populated clusters, with normalized entropy $0.978 \pm 0.007$ and a silhouette score of $0.154 \pm 0.013$ (ten seeds). The absolute silhouette is low, indicating overlapping rather than well-separated clusters, as expected for windows sampled along continuous trajectories. Cluster agreement with ten equal phase bins is moderate (NMI $0.265 \pm 0.013$), and a linear probe recovers the window's relative position in its episode from the embedding with $R^2 = 0.79 \pm 0.04$.

\begin{figure}[h]
\begin{center}
\includegraphics[width=0.9\linewidth]{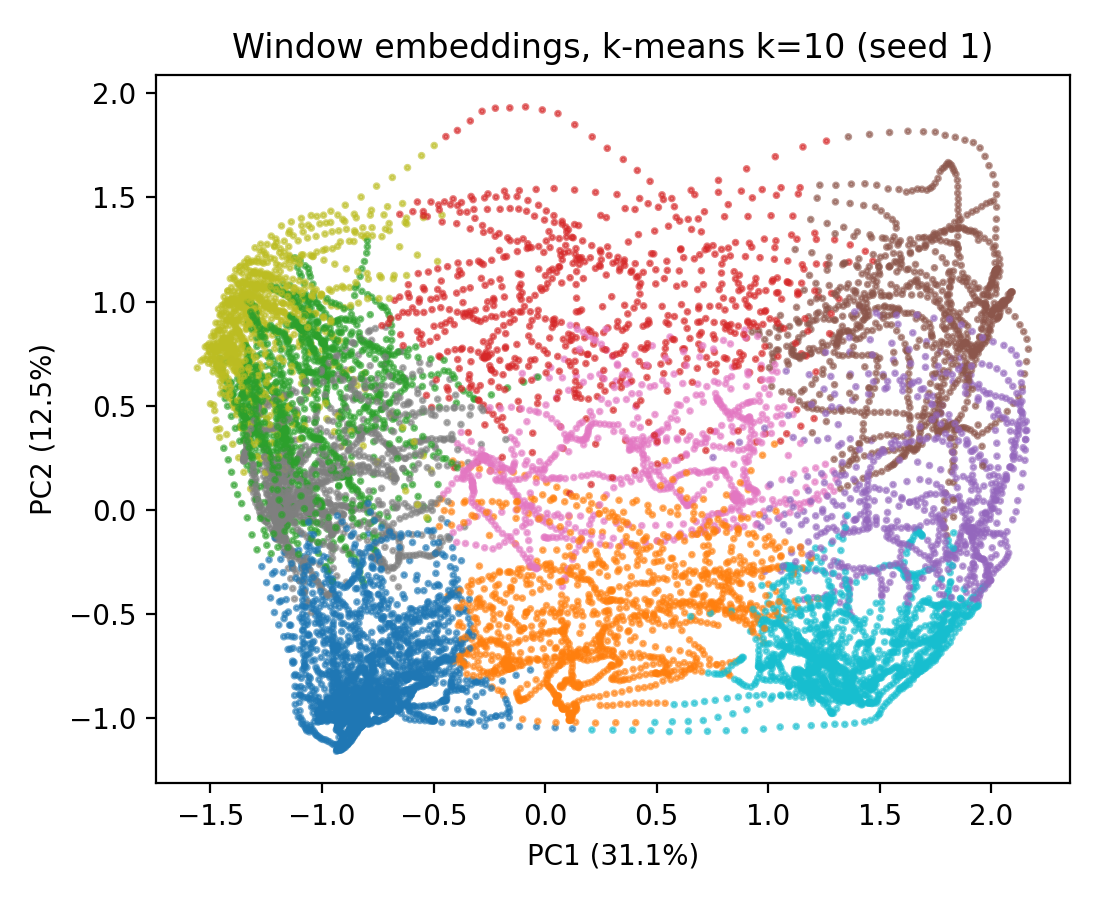}
\end{center}
\caption{
PCA of 64-dimensional RTDBN embeddings for 11{,}220 windows (seed 1), colored by $k$-means cluster ($k = 10$). PC1 and PC2 account for 31.1\% and 12.5\% of variance.
}
\label{fig:pca}
\end{figure}

A PCA projection of the embeddings, colored by cluster, is shown in Figure~\ref{fig:pca}.

\begin{figure}[h]
\begin{center}
\includegraphics[width=\linewidth]{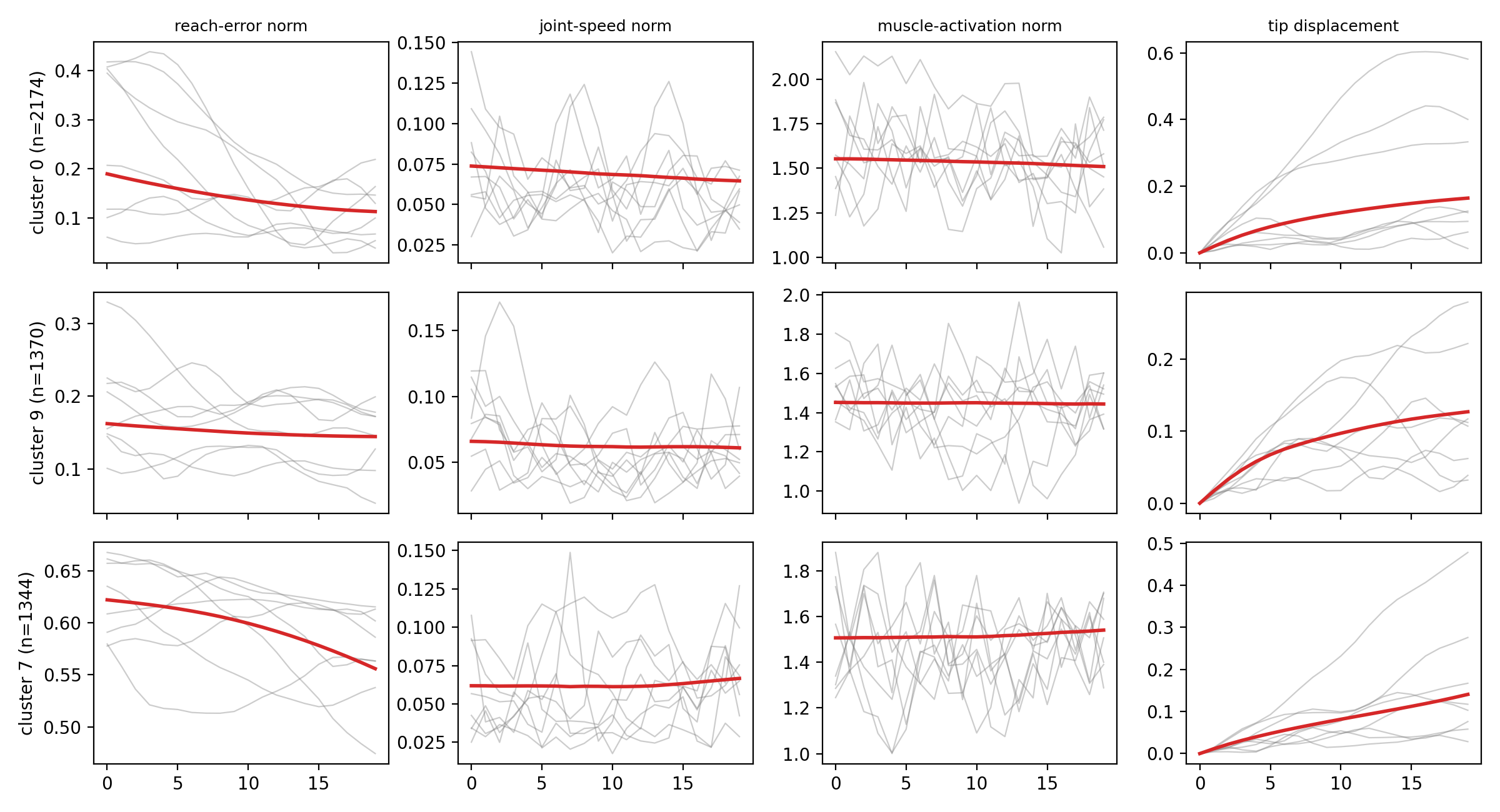}
\end{center}
\caption{
Eight randomly drawn example windows from each of the three largest clusters (rows; seed 1), cluster sizes 2{,}174, 1{,}370, and 1{,}344, across four observation-derived channels (columns): reach-error norm, joint-speed norm, muscle-activation norm, and tip displacement from the start of the window. Thin lines are individual windows, while the bold line is the per-cell mean.
}
\label{fig:cluster_examples}
\end{figure}

Looking at representative windows from the three largest clusters (seed 1) shows a kinematic reading organized around distance to the target and closing rate. The largest cluster (n = 2{,}174) is near the target (mean reach-error norm 0.143 m) but still closing it (falling by 0.077 m within the window, compared to a population mean reach-error norm of 0.358 m); the second (n = 1{,}370) is also near the target (0.151 m) and nearly static (falling by only 0.018 m, with the lowest mean muscle-activation norm of the three being 1.449); and the third (n = 1{,}344) is far from the target (0.597 m) and is closing slowly (0.066 m). (Figure~\ref{fig:cluster_examples}).

Across all 10 clusters, the mean reach-error norm falls into three bands (about 0.14--0.23 m, 0.43--0.50 m, and 0.58--0.65 m) and the closing rate varies within each band; the two fastest-closing clusters (n = 877 and n = 700) fall by 0.167 m and 0.157 m and have the highest joint speed (0.090) and tip displacement (0.328) of all ten. A more in-depth breakdown is provided in the appendix (Figure~\ref{fig:cluster_supplement}, Table~\ref{tab:cluster_all10}). These readings are merely descriptive as they are for one seed, and cluster identities and sizes differ across seeds.

\subsection{Test Split Generalization}

We apply the trained encoder and $k$-means centroids from the training split to the 15 held-out test episodes, which played no role in training. The split yields 1{,}980 windows. Held-out reconstruction MSE, computed with the same procedure on both sets, is $539 \pm 27$ on the test episodes and $547 \pm 29$ on the validation episodes, which is consistent with no overfitting to the training episodes. All 10 clusters stay populated on the test windows, the normalized entropy is $0.952 \pm 0.031$, and the silhouette score is $0.170 \pm 0.018$ on the test windows versus $0.154 \pm 0.013$ on train+validation. This indicates that the clustering is stable under an episode-disjoint split within our simulation setting.

\subsection{Ablation Study}

We vary $T \in \{10, 20, 30, 40\}$ (with $n_h = 64$ fixed) and $n_h \in \{32, 64, 128\}$ (with $T = 20$ fixed), measuring reconstruction MSE, silhouette score, and normalized entropy (Table~\ref{tab:ablation}).

\begin{table}[t]
\caption{Ablation results (five seeds, mean; learning rates selected at $T=20$, $n_h=64$ and not re-tuned). Test MSE is summed over the window ($\pm$ SD over seeds); MSE$/(Tn_v)$ divides it by $T \times 80$ to make window lengths comparable; $R^2$ is the median per-channel held-out value. All configurations produce 10 populated clusters.}
\label{tab:ablation}
\begin{center}
\begin{tabular}{lrrrrr}
\toprule
Configuration & Test MSE & MSE$/(Tn_v)$ & Median $R^2$ & Silhouette & $H_{\text{norm}}$ \\
\midrule
$T=10$, $n_h=64$ & 276 $\pm$ 14 & 0.345 & 0.610 & 0.148 & 0.976 \\
$T=20$, $n_h=64$ & 533 $\pm$ 26 & 0.333 & 0.610 & 0.153 & 0.978 \\
$T=30$, $n_h=64$ & 807 $\pm$ 65 & 0.336 & 0.602 & 0.143 & 0.969 \\
$T=40$, $n_h=64$ & 1034 $\pm$ 50 & 0.323 & 0.608 & 0.140 & 0.975 \\
\midrule
$T=20$, $n_h=32$ & 835 $\pm$ 49 & 0.522 & 0.549 & 0.212 & 0.974 \\
$T=20$, $n_h=64$ & 533 $\pm$ 26 & 0.333 & 0.610 & 0.153 & 0.978 \\
$T=20$, $n_h=128$ & 466 $\pm$ 64 & 0.291 & 0.678 & 0.141 & 0.977 \\
\bottomrule
\end{tabular}
\end{center}
\end{table}

Window length has almost no effect once reconstruction error is normalized by the window size: MSE$/(Tn_v)$ is 0.32--0.35 and median $R^2$ is 0.60--0.61 for every $T$, and silhouette varies only between 0.140 and 0.153 (within roughly one seed SD). Hidden size matters for reconstruction: median held-out $R^2$ rises from 0.549 at $n_h = 32$ to 0.610 at $n_h = 64$ and 0.678 at $n_h = 128$. Silhouette moves in the opposite direction (0.212, 0.153, 0.141).

\subsection{Determining if Recurrence Matters: A $\mathbf{W}'=0$ Ablation}
\label{sec:wprime}

The ablations above vary window length and hidden size, but do not test the paper's central premise directly: does the recurrent connection $\mathbf{W}'$ improve the representation? To explore this, we retrain the canonical $T=20$, $n_h=64$ configuration with $\mathbf{W}'$ initialized to zero and frozen throughout training, which removes the hidden state's dependence on its own history and reduces the model to a per-frame Gaussian RBM with learned variance, mean-pooled over the window as before.

\begin{table}[h]
\caption{$\mathbf{W}'=0$ ablation (ten seeds, mean $\pm$ SD; models paired by seed). Identical data, split, and training schedule to the canonical model; only $\mathbf{W}'$ is zeroed and frozen. Probe columns are test $R^2$ of a ridge probe on the embedding.}
\label{tab:wprime0}
\begin{center}
\begin{tabular}{lrr}
\toprule
 & Canonical ($\mathbf{W}'$ free) & $\mathbf{W}'=0$ \\
\midrule
Test MSE & $539 \pm 27$ & $538 \pm 31$ \\
Median channel $R^2$ (test) & $0.608 \pm 0.012$ & $0.615 \pm 0.018$ \\
Trainval silhouette & $0.154 \pm 0.013$ & $0.149 \pm 0.010$ \\
Test silhouette & $0.170 \pm 0.018$ & $0.174 \pm 0.018$ \\
Probe $R^2$: phase & $0.786 \pm 0.044$ & $0.779 \pm 0.054$ \\
Probe $R^2$: reach trend & $0.365 \pm 0.037$ & $0.376 \pm 0.043$ \\
Probe $R^2$: joint speed & $0.790 \pm 0.088$ & $0.825 \pm 0.057$ \\
\bottomrule
\end{tabular}
\end{center}
\end{table}

No metric in Table~\ref{tab:wprime0} differs between the two models: the paired difference in test MSE ($\mathbf{W}'=0$ minus canonical) is $-1.2$ (95\% bootstrap CI $[-19.0, +14.6]$, paired $t$ $p = 0.90$), in median $R^2$ $+0.006$ (CI $[-0.005, +0.017]$), and in test silhouette $+0.004$; no probe difference has a confidence interval excluding zero (joint speed: $+0.034$, CI $[-0.005, +0.087]$). Because the CIs are narrow, a benefit of $\mathbf{W}'$ larger than about 19 units of test MSE (3.5\%) or 0.005 in median $R^2$ can be excluded at this sample size.

\subsection{Number of Clusters}

\begin{figure}[h]
\begin{center}
\includegraphics[width=\linewidth]{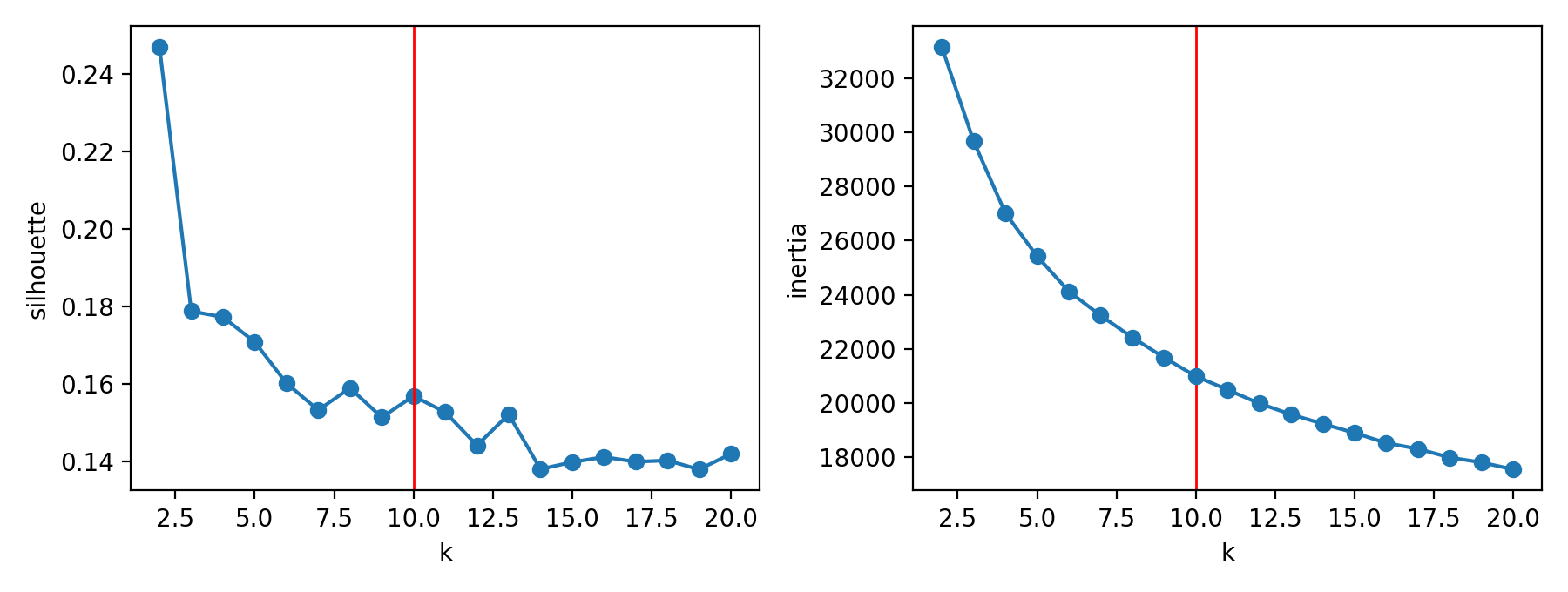}
\end{center}
\caption{Silhouette score (blue, left) and inertia (orange, right) versus $k$. Silhouette declines overall. Inertia decreases smoothly with no sharp elbow; the red line marks $k=10$, the value used throughout the rest of the paper.}
\label{fig:cluster_analysis}
\end{figure}

We sweep $k \in \{2, \ldots, 20\}$ to verify that $k = 10$ holds up and to look for degenerate behavior at high $k$ (Figure~\ref{fig:cluster_analysis}). Silhouette declines overall, from 0.247 at $k = 2$ down to 0.142 at $k = 20$ (seed 1) and is roughly flat at 0.14--0.16 for $k \geq 7$. The downward trend is expected when clusters are only weakly separated, as for windows sampled along continuous trajectories, because the silhouette score then tends to favor fewer, larger clusters \citep{teng2025silhouette}. The within-cluster sum of squares decreases from 33{,}148 to 17{,}546 across the full range with no sharp elbow at any single $k$; $k = 10$ (silhouette 0.157) is not distinguished by either curve, and it matches the fixed clustering resolution used throughout the rest of the paper's comparisons.

\subsection{Comparison with Non-RBM Baselines}
\label{sec:baselines}

We compared the RTRBM with plain non-probabilistic models on the same ten splits, scaling, windows, $k$-means, and probes: PCA on the flattened window or frame by frame, a frame-wise autoencoder, and a causal GRU autoencoder in which each frame is reconstructed from the recurrent state after seeing it (the closest non-probabilistic analogue of the RTRBM). All use 64-dimensional embeddings, the same 30 training epochs, and a learning rate from $\{10^{-3}, 3 \times 10^{-3}\}$ chosen on validation reconstruction error on one seed (Table~\ref{tab:baselines}).

\begin{table}[t]
\caption{RTRBM variants and non-RBM baselines on the held-out reach windows (ten seeds, mean $\pm$ s.d.). MSE is summed over the 20 frames and 80 channels of a window and averaged over windows (the RBM numbers use sampled binary hidden states; the baselines are deterministic). Silhouette is computed on the train and validation episodes ($k=10$). Probe columns are held-out $R^2$ of a ridge probe on the embedding.}
\label{tab:baselines}
\centering
\small
\resizebox{\linewidth}{!}{%
\begin{tabular}{lcccccc}
\toprule
Model & Test MSE & Median $R^2$ & Silhouette & Phase & Speed & Reach trend \\
\midrule
RTRBM, canonical & $539.5 \pm 26.8$ & $0.608 \pm 0.012$ & $0.154 \pm 0.013$ & $0.786 \pm 0.044$ & $0.790 \pm 0.088$ & $0.365 \pm 0.037$ \\
RTRBM, $\mathbf{W}'=0$ & $538.2 \pm 31.2$ & $0.615 \pm 0.018$ & $0.149 \pm 0.010$ & $0.779 \pm 0.054$ & $0.825 \pm 0.057$ & $0.376 \pm 0.043$ \\
\midrule
PCA, flattened windows & $478.8 \pm 14.4$ & $0.714 \pm 0.020$ & $0.088 \pm 0.002$ & $0.808 \pm 0.030$ & $0.329 \pm 0.079$ & $0.660 \pm 0.033$ \\
PCA, frame-wise & $0.0 \pm 0.0$ & $1.000 \pm 0.000$ & $0.098 \pm 0.003$ & $0.805 \pm 0.033$ & $0.345 \pm 0.069$ & $0.540 \pm 0.047$ \\
Frame-wise autoencoder & $2.3 \pm 0.5$ & $0.998 \pm 0.000$ & $0.077 \pm 0.007$ & $0.803 \pm 0.032$ & $0.327 \pm 0.081$ & $0.531 \pm 0.049$ \\
Causal GRU autoencoder & $2.0 \pm 0.4$ & $0.999 \pm 0.000$ & $0.080 \pm 0.008$ & $0.801 \pm 0.032$ & $0.326 \pm 0.081$ & $0.531 \pm 0.056$ \\
\bottomrule
\end{tabular}}
\end{table}

The RTRBM variants are not competitive reconstructors. Frame-wise PCA with 64 components reconstructs the held-out windows essentially exactly (median $R^2$ 1.000), because the 80 channels are highly redundant, and the autoencoders reach median $R^2$ of at least 0.998, against 0.608 for the canonical RTRBM and 0.615 for the $\mathbf{W}'=0$ variant. The baselines are better on the probes for the reach quantities (reach trend: 0.66 for windowed PCA against 0.37 for the canonical RTRBM; paired differences are positive for every baseline, Holm-adjusted $p < 0.001$). The RTRBM embeddings are better on the speed probe (mean absolute joint velocity: 0.79 against at most 0.35 for every baseline in Table~\ref{tab:baselines}). The $\mathbf{W}'=0$ model has the same speed probe (0.82), so this is not due to recurrence.

\subsection{Walking-Speed Classification on Two Gait Cohorts}
\label{sec:gait}

Table~\ref{tab:gait} gives the held-out-subject accuracies of the two probes. Among the learned encoders, the RTRBM embedding gives the highest random-forest accuracy in both cohorts: $0.577$ against $0.456$ to $0.515$ for the four learned baselines in cohort~1, and $0.610$ against $0.413$ to $0.559$ in cohort~2. All four contrasts hold after Holm correction in both cohorts (cohort~1: $+0.062$ to $+0.121$, Holm $p\le5.5\times10^{-4}$; cohort~2: $+0.051$ to $+0.197$, Holm $p\le0.018$). In the linear probes the RTRBM embeddings are better than frame-wise PCA, window PCA and the MLP autoencoder in cohort~2 (Holm $p=0.013$ to $0.033$) and than window PCA in cohort~1 (Holm $p=0.028$), and the remaining contrasts do not separate the encoders, so we do not claim an advantage there. With the $0.01\times$ multiplier on $\mathbf{W}'$ used for the reach data, the random-forest accuracies are $0.542$ (cohort~1) and $0.602$ (cohort~2), which is also above all four learned baselines.

\begin{table}[h]
\caption{Walking-speed accuracy on unseen subjects (mean over ten subject-level splits; chance $=0.333$). RTRBM: learned-variance Gaussian RTRBM; $\mathbf{W}'=0$ removes the recurrence; Stats: per-channel mean, standard deviation, minimum and maximum of the window (no training). Bold: best in row.}
\label{tab:gait}
\begin{center}
\small
\setlength{\tabcolsep}{4pt}
\begin{tabular}{llrrrrrrr}
\toprule
 & Probe & RTRBM & RTRBM $\mathbf{W}'=0$ & PCA frame & PCA window & MLP-AE & GRU-AE & Stats \\
\midrule
Cohort 1 & linear & 0.514 & 0.470 & 0.504 & 0.465 & 0.503 & 0.497 & \textbf{0.730} \\
(57 subj.) & random forest & 0.577 & 0.515 & 0.456 & 0.515 & 0.463 & 0.466 & \textbf{0.649} \\
\midrule
Cohort 2 & linear & 0.529 & 0.518 & 0.437 & 0.455 & 0.446 & 0.500 & \textbf{0.567} \\
(14 subj.) & random forest & 0.610 & 0.591 & 0.413 & 0.559 & 0.498 & 0.479 & \textbf{0.649} \\
\bottomrule
\end{tabular}
\end{center}
\end{table}

Removing the recurrence lowers random-forest accuracy by $0.062$ in cohort~1 (95\% CI $[+0.046, +0.078]$) but only by $0.019$ in cohort~2 (CI $[-0.017, +0.061]$), so recurrence appears to help in cohort~1 and is not distinguishable from zero in cohort~2. However, this claim is limited to these cohorts and task.

The training-free summary-statistics baseline is stronger than the RTRBM embedding. In cohort~1, it reaches $0.730$ with the linear probe and $0.649$ with the random forest, compared to $0.514$ and $0.577$ for the RTRBM (paired differences $+0.216$, 95\% CI $[+0.174, +0.258]$, and $+0.072$, CI $[+0.048, +0.094]$; Holm $p = 2\times10^{-5}$ and $7\times10^{-4}$ over the four contrasts). In cohort~2, it is numerically higher as well ($0.567$ and $0.649$ against $0.529$ and $0.610$), but the differences are within noise ($+0.038$, CI $[-0.008, +0.078]$, and $+0.039$, CI $[-0.009, +0.087]$; Holm $p = 0.28$). So the advantage over frame-wise and window PCA and over the autoencoders does not show that the RTRBM extracts walking-speed information that simple hand-crafted features lack. The summary statistics use $4\times$ as many dimensions as the 64-dimensional learned embeddings, and we ran this comparison after seeing the other results.

\section{Discussion}

Across ten seeds, multiple results emerged. First, learned-variance Gaussian visible units reconstruct held-out windows substantially better than Bernoulli and fixed-variance units (median $R^2$ 0.61 versus 0.09 and 0.57). Second, the clusters transfer to unseen episodes: the $k$-means centroids sort 15 held-out episodes into the same 10 groups with test silhouette at least as high as on the training episodes ($0.170 \pm 0.018$ versus $0.154 \pm 0.013$) and held-out reconstruction error as low as on validation episodes, which indicates that the clusters reflect properties of the reaching movement rather than of individual episodes.

One likely reason for the $\mathbf{W}'=0$ result is that reconstruction can be solved without memory; the canonical and $\mathbf{W}'=0$ models reach the same reconstruction error (\S\ref{sec:wprime}), so a reconstruction-based objective gives $\mathbf{W}'$ little to learn. We therefore do not interpret the null as evidence that recurrence cannot help, but instead as a property of reconstruction-based training on this dataset.

The data used in this study is intentionally limited in scope. Each episode uses a random action policy on a single fixed target, which gives us windows that are kinematically diverse but not behaviorally diverse. Meanwhile, variable-target reaching, task-directed policies, or real sensor data would give the encoder more to differentiate and would likely, as a result, produce richer cluster structure.

For future SIT-FUSE applications to sequential data, the results suggest the following practical use of the toolkit. The RTRBM/RTDBN embedding is a compact (64-dimensional) window descriptor that a nonlinear probe or a clustering stage can use directly, and on real gait data it is a better descriptor of walking speed than PCA or autoencoder embeddings under a random-forest probe, but not better than simple per-channel summary statistics (\S\ref{sec:gait}). The per-window outputs and TimeSHAP reports of the SIT-FUSE integration (\S\ref{sec:software}) then allow the resulting clusters to be inspected by channel and time step. In future work, larger and more varied cohorts and joint training with the information-theoretic clustering head are some natural next evaluations.

\section{Conclusion}

RTVarianceGaussianRBM, with its per-feature learned variance, a validation-selected learning rate and a two-phase warmup schedule, reconstructs held-out reaching windows far better than Bernoulli or fixed-variance alternatives (median $R^2$ 0.61 versus 0.09 and 0.57; ten seeds). The RTDBN collapses the resulting temporal hidden states into fixed-length embeddings whose $k$-means clusters carry a reading in terms of distance to the target and closing rate and transfer to held-out episodes. On two real gait cohorts the same embeddings classify walking speed in held-out subjects better than PCA and autoencoder embeddings under a random-forest probe, but not better than per-channel summary statistics, and the verified implementation runs end to end inside SIT-FUSE, from pretraining and clustering to per-window outputs and TimeSHAP attributions.

\subsubsection*{Broader Impact Statement}
This work develops unsupervised tools for analyzing biomechanical movement data. The simulated experiments use synthetic data, while the gait experiments use two publicly released datasets of healthy adults walking on a treadmill, with subjects identified only by code. The first dataset is released under a CC BY 4.0 license with written informed consent and institutional review board approval \citep{katmah2026multimodal}; the second is distributed through SimTK \citep{bailey2021simtk} with its associated publication \citep{bailey2021validity}. These methods make no clinical decisions, and any downstream use would need additional validation; we are also not aware of any significant negative impacts.

\bibliography{main}
\bibliographystyle{tmlr}

\appendix
\section{Implementation Details}

All models are implemented in PyTorch, extending the Learnergy library \citep{learnergy} and following its existing conventions. Source code for the temporal model classes, the dataset class, the experiments and the SIT-FUSE integration is provided as anonymized supplementary material, and will be released publicly upon acceptance.

Hyperparameters were selected by validation reconstruction MSE on one seed (seed 42, a split different from seeds 1--10); the grids were learning rate $\in \{0.001, 0.003, \ldots, 100\}$ (screening for 10 epochs at coarse spacing, then 30 epochs at $\{0.1, 0.3, 1, 3\}$ for the variance-Gaussian family), and the $\mathbf{W}'$ learning-rate multiplier $\in \{0.01, 0.1, 1\}$.

\section{Supplementary: All-Cluster Kinematic Breakdown}

\begin{figure}[h]
\begin{center}
\includegraphics[width=\linewidth]{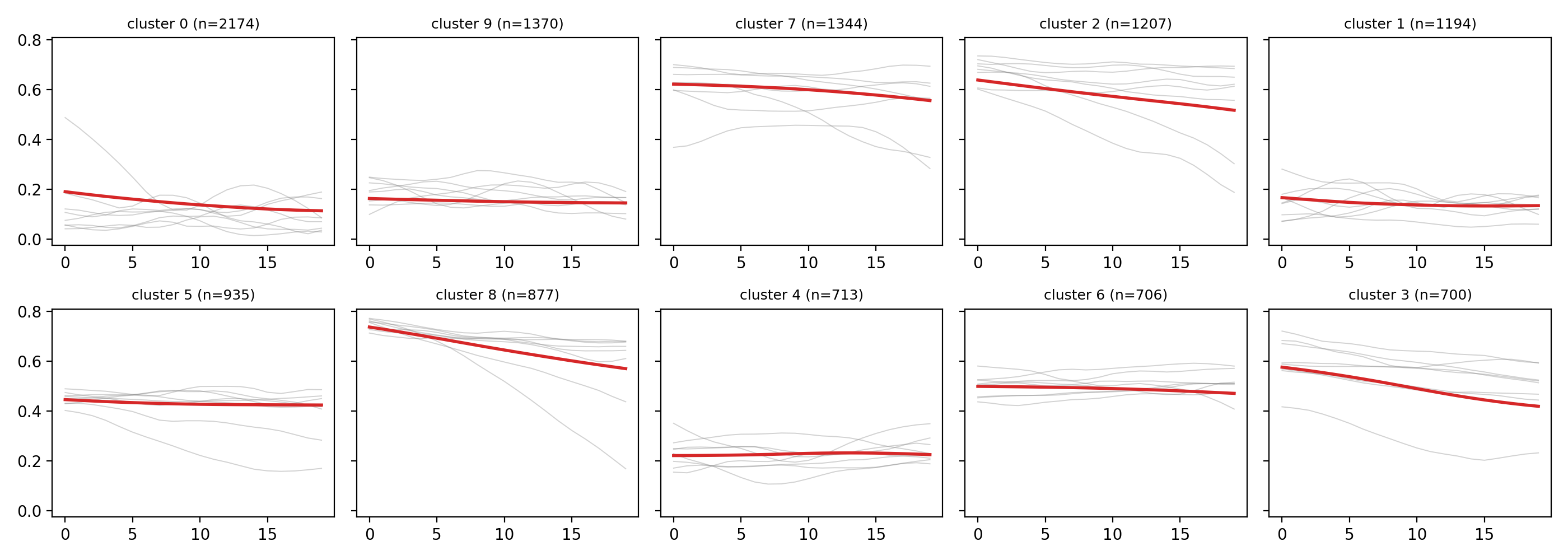}
\end{center}
\caption{
Reach-error magnitude by $k$-means cluster, all 10 clusters (seed 1; 8 example windows per cluster, thin lines; mean, bold line).
}
\label{fig:cluster_supplement}
\end{figure}

Figure~\ref{fig:cluster_supplement} and Table~\ref{tab:cluster_all10} extend Figure~\ref{fig:cluster_examples} to all 10 clusters, not only the three largest.

\begin{table}[H]
\caption{Per-cluster kinematic summary, all 10 clusters. Trend is the change in reach-error norm from the start to the end of the window (negative = closing distance to target). Joint speed, muscle effort, and tip displacement are window-mean norms.}
\label{tab:cluster_all10}
\begin{center}
\begin{tabular}{lrrrrrr}
\toprule
Cluster & $n$ & Reach err. & Trend & Joint spd. & Muscle eff. & Tip disp. \\
\midrule
0 & 2{,}174 & 0.143 & $-$0.077 & 0.069 & 1.535 & 0.164 \\
1 & 1{,}194 & 0.142 & $-$0.032 & 0.071 & 1.502 & 0.172 \\
2 & 1{,}207 & 0.577 & $-$0.122 & 0.082 & 1.555 & 0.224 \\
3 & 700 & 0.496 & $-$0.157 & 0.084 & 1.666 & 0.328 \\
4 & 713 & 0.226 & $+$0.004 & 0.076 & 1.500 & 0.202 \\
5 & 935 & 0.430 & $-$0.022 & 0.070 & 1.570 & 0.205 \\
6 & 706 & 0.488 & $-$0.028 & 0.066 & 1.525 & 0.132 \\
7 & 1{,}344 & 0.597 & $-$0.066 & 0.063 & 1.517 & 0.141 \\
8 & 877 & 0.652 & $-$0.167 & 0.090 & 1.512 & 0.265 \\
9 & 1{,}370 & 0.151 & $-$0.018 & 0.063 & 1.449 & 0.127 \\

\bottomrule
\end{tabular}
\end{center}
\end{table}

\end{document}